\documentclass[nohyperref]{article}

\usepackage{microtype}
\usepackage{graphicx}
\usepackage{subfigure}
\usepackage{booktabs} 

\usepackage{hyperref}

\usepackage[accepted]{icml2022}

\usepackage{amsmath}
\usepackage{amssymb}
\usepackage{mathtools}
\usepackage{amsthm}

\usepackage[capitalize,noabbrev]{cleveref}

\theoremstyle{plain}

\theoremstyle{definition}

\theoremstyle{remark}

\usepackage[textsize=tiny]{todonotes}

\icmltitlerunning{Symplectically Integrated Symbolic Regression}

\begin{document}

\twocolumn[
\icmltitle{Symplectically Integrated Symbolic Regression\\
          of Hamiltonian Dynamical Systems}



\icmlsetsymbol{equal}{*}

\begin{icmlauthorlist}
\icmlauthor{Daniel M. DiPietro}{}
\icmlauthor{Bo Zhu}{yyy}
\end{icmlauthorlist}

\icmlaffiliation{yyy}{Department of Computer Science, Dartmouth College, Hanover NH.}

\icmlcorrespondingauthor{Daniel DiPietro}{dipietrodaniel131@gmail.com}

\icmlkeywords{Machine Learning, Computational Physics}

\vskip 0.3in
]



\printAffiliationsAndNotice{}  

\begin{abstract}
Here we present Symplectically Integrated Symbolic Regression (SISR), a novel technique for learning physical governing equations from data. SISR employs a deep symbolic regression approach, using a multi-layer LSTM-RNN with mutation to probabilistically sample Hamiltonian symbolic expressions. Using symplectic neural networks, we develop a model-agnostic approach for extracting meaningful physical priors from the data that can be imposed on-the-fly into the RNN output, limiting its search space. Hamiltonians generated by the RNN are optimized and assessed using a fourth-order symplectic integration scheme; prediction performance is used to train the LSTM-RNN to generate increasingly better functions via a risk-seeking policy gradients approach. Employing these techniques, we extract correct governing equations from oscillator, pendulum, two-body, and three-body gravitational systems with noisy and extremely small datasets.

\end{abstract}

\section{Introduction}

Newton, Kepler, and many other physicists are remembered for uncovering mathematical governing equations of complex physical dynamical systems. Ultimately, they were performing exceedingly difficult free-form regression tasks, finding relationships in data sourced from real-world physical observations.

Modern machine learning techniques have demonstrated a broad and impressive ability to extract meaningful relationships from data \cite{gpt2, classification1, classification2}. Researchers are increasingly interested in applying such techniques to scientific discovery tasks, particularly in the context of physical dynamical systems and their governing equations \cite{pdenet, pdenet2, datadriven, fluids, karniadakis, sindy, tong2021symplectic, poincare}. When compared to traditional machine learning tasks, governing equation discovery offers additional changes in that observation datasets tend to be relatively small and are inevitably tainted with noise or measurement error. Constructing approaches that can learn from a small number of observations while being resistant to noise is a core goal of using machine learning for governing equation discovery. Much of this work has focused on specific formalisms of governing equations, such as Hamiltonians, which provide a consistent approach for many physical dynamical systems and the easy incorporation of priors to shrink the solution space \cite{hnn, ssinn, srnn}.

In the past literature, there have been two lines of approaches to learn and predict the time evolution of a physical dynamical system. The first approach emphasizes prediction, relying upon black-box machine learning techniques that can time-evolve the dynamical system with very high accuracy but offer no insight into its underlying governing equations \cite{hnn, srnn, raissi2020hidden, newtonvsmachine, lagrangian}. The second approach emphasizes interpretability, prioritizing the discovery of true, concise governing equations for the system  \cite{ssinn, sindy, genetic, gradient2, sparse3, poincare}. Interpretable approaches have yielded noteworthy successes, but they often require large amounts of data, fail to learn systems with many degrees of freedom, or require the user to specify function spaces that the governing equation must be contained in. Both black-box and interpretable methods commonly incorporate physics priors, such as symplectic integration schemes, modified loss functions, structure-enforcing formalisms, and constraining equations derived from known physical laws.

In this paper, we present a novel interpretable method: Symplectically Integrated Symbolic Regression (SISR). SISR adopts the Hamiltonian formalism, seeking to uncover governing Hamiltonians of physical dynamical systems solely from position-momenta time series data. First, SISR employs a model-agnostic technique using black-box symplectic networks to extract common Hamiltonian priors, such as coupling, from data. Next, SISR uses the deep symbolic regression paradigm, employing an autoregressive LSTM-RNN that probabilistically generates sequences of function operators, variables, and placeholder constants. These sequences correspond to pre-order traversals of free-form symbolic expressions: in this case, Hamiltonians. The RNN is specially constructed so that only separable Hamiltonian functions are generated. Additionally, the Hamiltonian priors extracted in the initial step are imposed into RNN architecture so that all generated Hamiltonians obey them, drastically reducing the search space. Once generated, these Hamiltonian sequences are translated at run-time into trainable expressions that support auto-differentiation. Each expression has its constants optimized via back-propagation through a symplectic integrator; its prediction performance (reward) is recorded during this process. Then, using a risk-seeking policy gradients approach, the RNN is trained to generate increasingly better symbolic expressions.

Using these techniques, SISR manages to extract correct governing Hamiltonians from four noteworthy dynamical systems, including harmonic oscillator, pendulum, two-body gravitational, and three-body gravitational. All experiments are performed on small training datasets (none them exceed 200 points, as compared to 1000+ in other state-of-the-art approaches), with and without Gaussian noise. To our knowledge, SISR is the first free-form, interpretable machine learning approach that can discover governing Hamiltonians of systems with many degrees of freedom using small-scale, noisy data samples.

In summary, we propose SISR, a novel technique for extracting governing Hamiltonians from observation data, which makes the following key architectural contributions:
\begin{enumerate}
    \item Employs a model-agnostic approach using symplectic neural networks to extract common Hamiltonian priors from data.
    \item Uses a specially constructed LSTM-RNN with mutation to generate separable Hamiltonians that obey the extracted Hamiltonian priors.
    \item Imposes a further symplectic prior on the RNN-generated expressions by training their constants via backpropagation through a symplectic integrator.
\end{enumerate}

\section{Related Work}

\paragraph{Symbolic Regression: Traditional and Deep} Often, regression techniques parameterize underlying functions using specific functional forms, such as lines or polynomials. Symbolic regression, on the other hand, performs its regression search over the space of all mathematical functions that can be constructed from some given set of operators: the form is essentially arbitrary. Traditionally, symbolic regression has relied upon genetic programming techniques \cite{fieldguide, genetic2, genetic}. Recently, researchers have begun incorporating modern deep learning approaches. \citet{sahoo2018learning} parameterize spaces of symbolic functions using neural networks with specially constructed activation functions. More recently, \citet{deepsymbolicregression} introduce the deep symbolic regression model, which uses an autoregressive RNN to generate symbolic expressions and then trains it using a novel policy-gradients approach. \citet{deepsymbolicregressionex2} augment the deep symbolic regression approach, using an RNN-guided symbolic expression search to construct populations for use with genetic programming. Note that most of these techniques are for the general symbolic regression task, lacking any priors unique to physical dynamical systems.

\paragraph{Computational Techniques for Governing Equation Discovery} There has been considerable research done on machine learning models that can extract mathematically concise governing functions from physical systems. \citet{sindy} propose SINDy, which operates within a pre-specified function space (i.e. polynomial, trigonometric, etc.) and performs a L1-regularized sparse regression, learning which terms to keep. \citet{ssinn} introduce SSINNs, which use a similar approach to SINDy but frame the problem using the Hamiltonian formalism and perform the sparse optimization by back-propagating through a symplectic integrator, which incorporates a symplectic bias into the symbolic expressions. These sparse approaches are limited in that the governing function must exist in the pre-specified function space in order to be found. \citet{genetic} use graph neural networks, representing particles as nodes and forces between pairs of particles as edges. After training these graph models, they extract symbolic expressions by using Eureqa to fit the edge, node, and global models of their graph networks. However, their experiments involve a relatively large volume of clean data. \citet{gradient2} employ a genetic, free-form symbolic regression approach that focuses upon conserved quantities such as Lagrangians and Hamiltonians; they succeed in learning governing equations for systems such as the double pendulum. Finally, there are numerous other approaches for equation discovery that follow different paradigms altogether \cite{pdenet, pdenet2, karniadakis, saemundsson2020variational, sun2019neupde}.

\paragraph{Symplectic Machine Learning Models} In recent years, many works have incorporated symplectomorphism as a physics prior; as a result, they tend to use the Hamiltonian formalism as well. \citet{hnn} propose Hamiltonian Neural Networks, which parameterize Hamiltonians using black-box deep neural networks. They optimize the gradients of these networks via back-propagation through a leapfrog numerical integrator, which gives rise to conserved quantities in the model. \citet{srnn} employ a similar approach but opt for a symplectic integrator and recurrent neural network; they achieve noteworthy prediction results but, due to their choice of integrator, can only learn separable Hamiltonians. \citet{xiong2020nonseparable} provide a novel architecture that incorporates symplectic priors but is not limited to separable Hamiltonians; they achieve noteworthy results on both separable and non-separable systems, including chaotic vortical flows. \citet{ssinn}, mentioned above, use a symplectic prior to augment an interpretable sparse regression approach. Symplectic priors are used by a variety of other works as well \cite{tong2021symplectic, zhu2020deep, mattheakis2019physical, jin2020sympnets}. 

\section{Background}

SISR learns governing equations in the Hamiltonian formalism and employs a model-agnostic approach to extract Hamiltonian coupling priors. Here we offer mathematical background on Hamiltonian mechanics and coupling.

\subsection{Hamiltonian Mechanics}

Hamiltonian mechanics is a reformulation of classical mechanics introduced in 1833 by William Rowland Hamilton. The Hamiltonian formalism provides a consistent approach for mathematically describing a variety of physical dynamical systems; it is heavily used in quantum mechanics and statistical mechanics, often for systems with many degrees of freedom \cite{introclassical, statphysics, classmechanics, quantmechanics, analyticalmechanics}.

In Hamiltonian mechanics, dynamical systems are governed via their Hamiltonian \(\mathcal{H}(\mathbf{p}, \mathbf{q})\), a function of the momenta and position vectors (\(\mathbf{q}\) and \(\mathbf{p}\) respectively) of each object in the system. The value of the Hamiltonian generally measures the total energy of the system; for systems where energy is conserved, \(\mathcal{H}\) is time-invariant. Further, we say that \(\mathcal{H}\) is \textit{separable} if it can be decomposed into two additively separable functions \(T(\mathbf{p})\) and \(V(\mathbf{q})\), where \(T\) measures kinetic energy and \(V\) measures potential energy. 

The time evolution of a physical dynamical system is uniquely defined by the partial derivatives of its Hamiltonian:
\begin{equation}
    \frac{d \mathbf{p}}{dt} = -\frac{\partial \mathcal{H}}{\partial \mathbf{q}}, \frac{d \mathbf{q}}{dt} = \frac{\partial \mathcal{H}}{\partial \mathbf{p}}
\end{equation}
The time evolution of a Hamiltonian system is symplectomorphic, preserving the symplectic two-form \(d\mathbf{p} \land d\mathbf{q}\). As a result, researchers generally time-evolve Hamiltonians with symplectic integrators, which are designed to preserve this two-form. We employ a fourth-order symplectic integrator (pseudocode in Appendix \ref{app:symp}) \cite{fourthorder}.

\subsection{Coupling}\label{sec:coupling}

Hamiltonians commonly exhibit \textit{coupling}: they can be written as a series of additively separable functions where each function takes some smaller subset of variables. Consider some arbitrary n-body, m-dimensional separable Hamiltonian $\mathcal{H}(\mathbf{p}, \mathbf{q}) = T(\mathbf{p}) + V(\mathbf{q})$ where the bodies are labeled $1,\dots,n$ and the dimensions are labeled $1,\dots,m$. What coupling might occur?

Without loss of generality, consider $T(\mathbf{p})$. Denote the momentum of body $i$ in dimension $j$ as $p_{i,j}$. We say that $T(\mathbf{p})$ exhibits complete decoupling if it may be rewritten as the sum of functions that each take only one variable in the system, i.e. it may be written in the form:
\begin{equation}\label{eq:none}
\begin{split}
    T(\mathbf{p}) = f_1(p_{1,1}) + \dots + f_m(p_{1,m}) + \dots +\\ f_{nm-m}(p_{n,1}) + \dots + f_{nm}(p_{n,m})
\end{split}
\end{equation}
We say that $T(\mathbf{p})$ exhibits dimensional coupling if it may be rewritten as the sum of functions that each take all dimensional variables for only a single particle in the system, i.e. it may be written in the form:
\begin{equation}\label{eq:dimension}
\begin{split}
    T(\mathbf{p}) = g_1(p_{1,1},\dots,p_{1,m}) + \dots +\\ g_{m}(p_{n,1},\dots,p_{n,m})
\end{split}
\end{equation}
Finally, we say that $T(\mathbf{p})$ exhibits pairwise coupling if it may be rewritten as the sum of functions that each take all dimensional variables for a pair of particles in the system, i.e. where $\mathbf{p_i}$ denotes all momenta variables of particle $i$, $T$ may be written in the form:
\begin{equation}\label{eq:pairwise}
\begin{split}
    T(\mathbf{p}) = h_1(\mathbf{p}_1,\mathbf{p}_2) + \dots + h_{m-1}(\mathbf{p}_1,\mathbf{p}_{n}) +\\
    h_m(\mathbf{p}_2, \mathbf{p}_3) + \dots + h_{2m-2}(\mathbf{p}_2, \mathbf{p}_n) + \dots
\end{split}
\end{equation}

\subsection{Coupling Composite Functions}\label{sec:composite}

Consider some arbitrary coupling function $f(x_1, \dots, x_n)$. Often, $f$ may be written as a composition of functions $f(x_1, \dots, x_n) = g(h(x_1, \dots, x_n))$ where $h$ performs one of several common reduction operations, such as summing, multiplication, Euclidean distance, or Manhattan distance. In these instances, we would say that particles are coupled via the given inner operation, such as their sum. If $h$ is known, then one may learn $f$ by learning $g$, which tends to be much simpler than learning the entire $f$ directly.

\subsection{Symmetries}\label{sec:symmetries}

Suppose that a Hamiltonian function exhibits pairwise coupling, i.e., it may be written as the sum of many additively separable coupling functions. Often, across all pairs of particles, these coupling functions are identical (symmetric). For instance, in an n-body gravitational system with particles of equal mass, or an n-body electrostatic system with particles of equal charge, all particles are pairwise coupled using the exact same coupling function. 

\section{Algorithm}

Here we present the SISR architecture, which consists of a coupling extraction process followed by a deep symbolic regression-inspired training loop. This training loop can further be decomposed into expression generation (\ref{sec:expgen}), symplectic optimization (\ref{sec:sympopt}), and RNN training (\ref{sec:rnntrain}).

\subsection{Extracting Function Properties via Symplectic Neural Networks}\label{sec:sympnets}

As detailed in Sections \ref{sec:coupling}, \ref{sec:composite}, and \ref{sec:symmetries}, Hamiltonians commonly exhibit coupling properties. These properties may be used to simplify the Hamiltonian discovery process by imposing restrictions that dramatically reduce the solution space. Thus, the first step of SISR is to find what coupling is taking place in the Hamiltonian, whether it can be reduced into simpler composite functions, and whether it is symmetric across particles or pairs of particles.

Our approach for doing this is rooted in symplectic neural networks as effective black-box predictors of Hamiltonian dynamical systems. Specifically, we can construct multi-layer neural networks that enforce specific coupling properties. Then, we may train these networks by using them to time-evolve the dynamical system through a symplectic integrator. By seeing how these networks perform with and without imposed coupling properties, we may deduce whether those properties hold for the system. To the best of our knowledge, we are the first work to extract coupling properties in the context of Hamiltonian dynamical systems, and we are not aware of any other work that employs symplectic neural networks in a similar manner.

\subsubsection{Coupling}

\begin{figure}[ht]
\vskip 0.2in
\begin{center}
\centerline{\includegraphics[width=0.80\columnwidth]{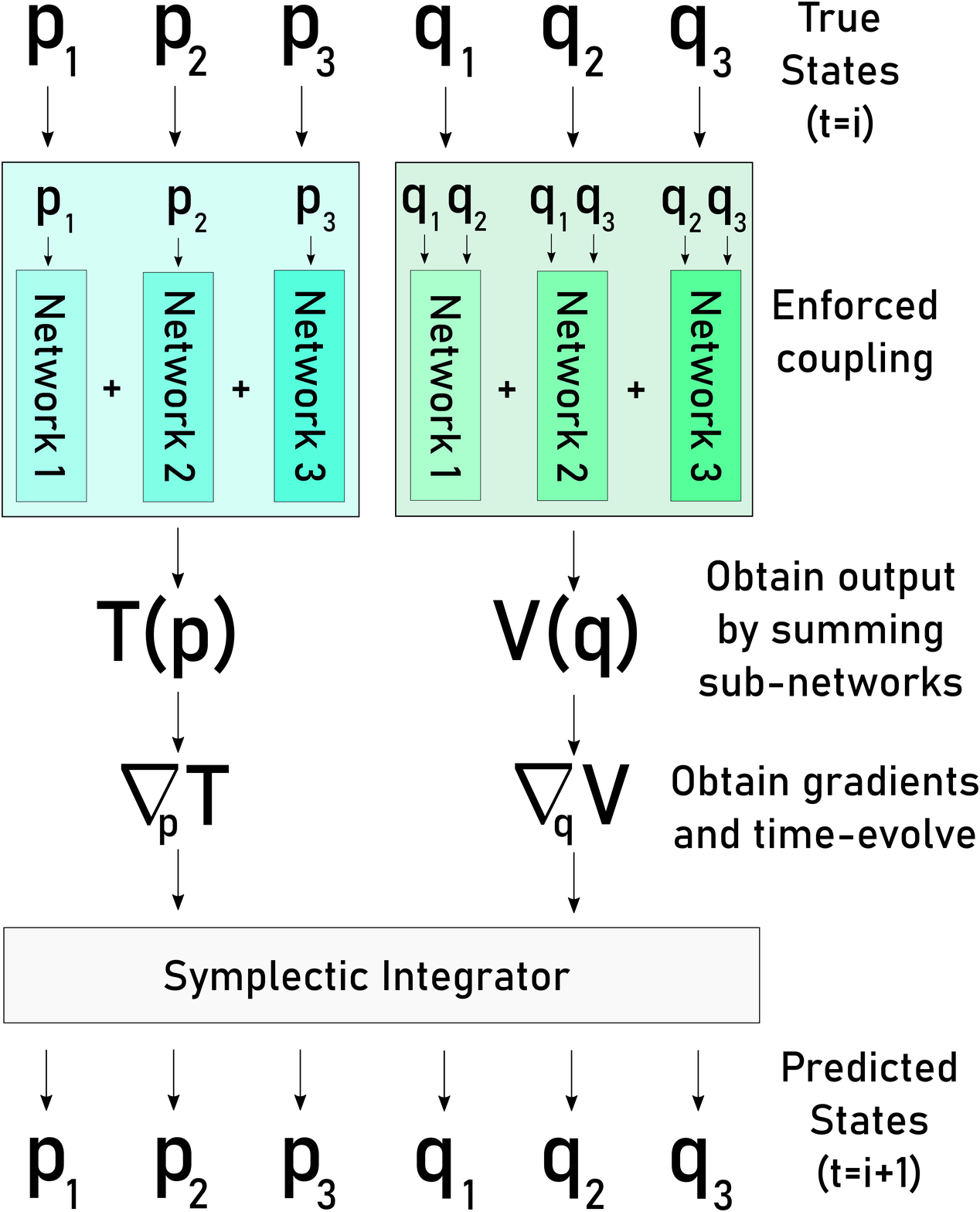}}
\caption{Symplectic network constructed with 3 particles exhibiting dimensional position coupling and pairwise momentum coupling. Once the predicted states at $t=i+1$ are obtained, the model can be trained by taking loss with respect to the actual states at $t=i+1$.}
\label{fig:pairwisemodel}
\end{center}
\vskip -0.2in
\end{figure}

In order to assess whether a physical dynamical system exhibits a particular variety of coupling, we can construct a model that enforces one of the functional forms described in Equations \ref{eq:none}, \ref{eq:dimension}, or \ref{eq:pairwise}. We do this by parameterizing the individual coupling functions with neural networks and summing the individual outputs of these networks to get the entire function output for $V$ or $T$. Note that these models are trained via gradient descent through a symplectic integrator; this process is depicted for a 3-body pairwise-momentum dimensional-position model in Figure \ref{fig:pairwisemodel}.

Although this method works for assessing specific instances of coupling, the question still remains of how we construct a search over $T$ and $V$, as the couplings of these functions are independent. Additionally, even if a Hamiltonian exhibits pairwise coupling, it doesn't mean that \textit{all} particles in the system are necessarily pairwise coupled. Our search approach is simple and works well empirically.

We begin by training a model where $T$ and $V$ are each parameterized by a single network, meaning that there are no coupling properties enforced. The value of the performance metric for this model is used to set a baseline. Then, holding this network constant for $T$, we enforce complete decoupling in $V$. If the corresponding performance metric drop is above some pre-defined fixed percentage relative to the baseline, denoted the \textit{maximum tolerable performance decrease}, we say that the $V$ is in fact completely decoupled. Otherwise, we attempt dimensional coupling and possibly pairwise coupling. This process is then repeated for $T$, with the network in $V$ held constant. Note that, if pairwise coupling is found, we perform a recursive backwards elimination to see which pairs must be present; this elimination process requires an additional maximum tolerable performance decrease parameter; this parameter is generally smaller, as it is easy to overfit to pairs that aren't actually present in the Hamiltonian, especially with little data.

\begin{figure*}[ht]
\vskip 0.2in
\begin{center}
\centerline{\includegraphics[width=2\columnwidth]{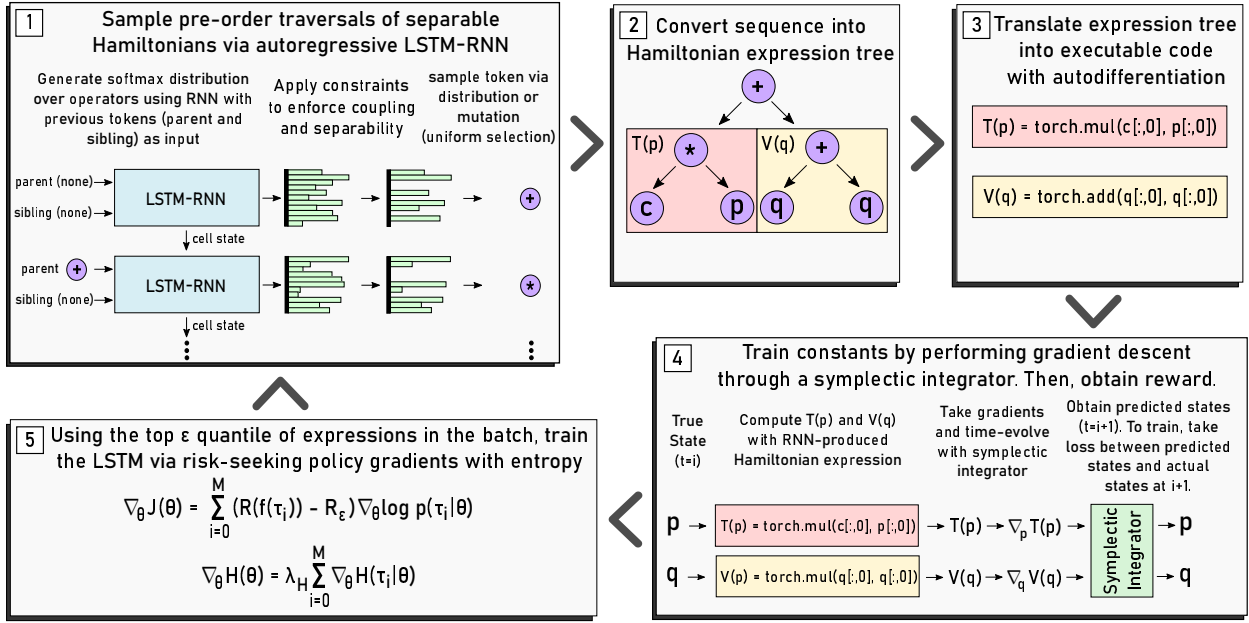}}
\caption{Overview of the SISR Hamiltonian generation, symplectic optimization, and RNN training processes. Note that steps 1-4 occur in batches, with many variable-length expressions being sampled, converted, translated, and trained in parallel.}
\label{fig:modeloverview}
\end{center}
\vskip -0.2in
\end{figure*}

\subsubsection{Coupling Composite Functions}

Once coupling has been identified, we determine whether the quantities are coupled by their sum, product, distance, or some other operation. To do this, we modify the network so that the operation being tested is performed immediately on the input variables prior to the forward pass. Then, we select the best performing composite function that meets the maximum tolerable performance decrease.

\subsubsection{Symmetries}

Once we have determined which particles are coupled and by what, if any, composite functions, we assess whether the coupling functions are symmetric across all pairs of particles. To do this, we use the same neural network for all coupled quantities (such as all pairs of positions in an n-body gravitational system).  Note that this requires that the domain of input values for each particle are reasonably close. If the performance doesn't degrade below our tolerance, we say that the couplings are symmetric.

\subsection{Generating Symbolic Hamiltonians}\label{sec:expgen}

With our coupling properties determined, we are ready to begin generating candidate symbolic Hamiltonians. We do this via a deep symbolic regression approach, employing an LSTM-RNN.

First, note that free-form symbolic functions may be represented as trees where internal nodes are functional operators and leaves are variables or constants. The pre-order traversal of this expression tree produces a sequence that uniquely defines the symbolic expression. Thus, sequence-generating RNNs can be used in an autoregressive manner to produce symbolic functions. We refer to functional operators, variables, and constants as \textit{tokens} in the sequence.

To generate a symbolic expression, an LSTM-RNN generates a sequence from left-to-right. As input, the network takes a one-hot encoding of the parent and sibling token for the current token being sampled. It then emits a softmax distribution, which is adjusted to eliminate all invalid tokens. This distribution may be sampled to generate the next token in the sequence. We also employ an epsilon-greedy mutation approach where the next token is sampled uniformly from the list of valid tokens with $\mu$ probability, where $\mu$ is a pre-defined hyperparameter. This process is depicted in Step 1 of Figure \ref{fig:modeloverview}.

Due to our use of a fourth-order symplectic integrator, we require that any Hamiltonian generated by the LSTM-RNN is of the form $\mathcal{H}(\mathbf{p}, \mathbf{q}) = T(\mathbf{p}) + V(\mathbf{q})$. This constraint is applied \textit{in situ}, meaning that the softmax distribution over the operators is constantly adjusted as expressions are generated so that only Hamiltonians of this form will ever be produced. This same approach is used to incorporate any coupling properties previously discovered: the emitted distribution is adjusted prior to sampling a token so that we will only ever see Hamiltonians that exhibit the desired coupling. To the best of our knowledge, this is the first instance of an LSTM-RNN being used to generate separable Hamiltonians with desired coupling properties.

Once a Hamiltonian sequence is generated, it is translated into an executable model with auto-differentiation (Steps 2 and 3 of Figure \ref{fig:modeloverview}). The first reason for this is to that constants can be optimized--the sequence contains placeholder ``ephemeral'' constants that do not refer to specific values. The second purpose is so that the performance of the expression can be assessed. This is done by numerically integrating the Hamiltonian to predict the system, which requires the ability to take its gradients.

\subsection{Symplectic Optimization}\label{sec:sympopt}

As mentioned previously, candidate Hamiltonian functions must have their placeholder constants trained. This is performed via a symplectic optimization approach (Step 4 of Figure \ref{fig:modeloverview}). Using a fourth-order symplectic integrator and a time-series of observation data for the system, we use the candidate Hamiltonian to time-evolve the system and obtain predicted future states. We then take the loss between our predicted future states and the actual future states contained in the observation data, backpropagating through our integrator to train the constants and thus imparting an energy-conserving symplectic prior onto any Hamiltonian generated by the RNN. This approach, which has been applied to black-box system prediction methods with great success, has yet to be employed for any free-form symbolic regression model \cite{srnn, ssinn}.

\subsection{Training the RNN}\label{sec:rnntrain}

Thus far, we have presented an RNN that can generate separable Hamiltonians (with specific coupling properties) that are then trained via backpropagation through a symplectic integrator. However, in order for a symbolic Hamiltonian search to succeed, the RNN must be trained to produce progressively better-fitting Hamiltonians.

To do this, we employ the risk-seeking policy gradients approach presented by \citet{deepsymbolicregression}, which maximizes the best-case expectation of rewards for expressions generated by the policy (in this case the RNN). To do this, we train on the top $\epsilon$ quantile in some given batch, where $1-\epsilon$ is referred to as the \textit{risk factor}. First, let this quantile consist of $M$ expressions where $\tau_{i}$ denotes the $i$-th expression in the quantile. If $f$ defines a symplectic integration function, $R$ defines a reward function, $R_\epsilon$ defines the reward of the top $\epsilon$ quantile, and $J(\theta)$ defines the expectation of rewards in the top $\epsilon$ quantile conditioned on the weights $\theta$ of the RNN, we have:
\begin{equation}
    \nabla_\theta J(\theta) = \sum_{i=1}^M \left(R(f(\tau_i)) - R_\epsilon\right) \nabla_\theta \log p(\tau_i | \theta)
\end{equation}

In the event that the expression generation involved mutation, we compute $\log p(\tau_i | \theta)$ using the probabilities produced by the RNN--\textit{not} the uniform probabilities generated to sample the mutated operator.

Further, as in \citet{deepsymbolicregression}, we adopt the maximum entropy framework and include an entropy term in the loss \cite{entropy}. Where $\lambda_H$ is the entropy coefficient, we have:
\begin{equation}
    \nabla_\theta H(\theta) = \lambda_H \sum_{i=1}^M \nabla_\theta H(\tau_i | \theta)
\end{equation}
Note that $H$ refers to the Shannon entropy function and is \textit{not} a Hamiltonian.

This training process corresponds to Step 5 of Figure \ref{fig:modeloverview}. We are not aware of any other approach for Hamiltonian dynamical systems that employs these reinforcement learning-oriented training approaches.

\section{Experiments}

We assess SISR by applying it to four noteworthy Hamiltonian dynamical systems: the harmonic oscillator, the pendulum, the two-body gravitational system (the Kepler problem), and the three-body gravitational system. Aside from expression length, all experiments use the same set of hyperparameters, which are almost entirely replicated from \citet{deepsymbolicregression} and detailed in Appendix \ref{app:hyper}.

For each experiment, we generated 3 different time series datasets via fourth-order symplectic integration with different initial conditions, which are available in Appendix \ref{app:initial}. These datasets are exceedingly small--smaller than any other work in this domain that we are aware of. Each experiment was conducted a total of 5 times on each dataset. Note that each dataset consists entirely of training data, as either SISR extracts the correct governing Hamiltonian from the training data or it does not. We say that a SISR-generated Hamiltonian is correct if it is algebraically equivalent to the true Hamiltonian with $10^{-2}$ precision in its constants. Our LSTM-RNN was optimized with Adam and employed an NRMSE reward function.

In initial experiments, we noticed very high variance in the number of epochs needed to obtain the Hamiltonian. We believe that this may be a consequence of the initial batch being too small to adequately explore the function space. As a result, we set the initial batch size to $4$ times the regular batch size; we adjust the risk factor to compensate, meaning that the RNN is trained on the same number of expressions after each batch.

\subsection{Harmonic Oscillator}

The harmonic oscillator is a one-dimensional, single-particle dynamical system described by the Hamiltonian
\begin{equation}
    \mathcal{H}(p,q) = \frac{p^2}{2m} + \frac{1}{2}m \omega^2 q^2
\end{equation}
where $m$ is the mass of the particle and $\omega$ refers to its oscillator frequency.

We trained SISR on 3 different datasets of harmonic oscillator motion with varying initial conditions, each consisting of only $30$ points from $t=0$ to $t=3$. In addition to variables and constants, SISR had access to the operators $\{+,-,\div,\cdot,\wedge\}$. The $T$ and $V$ of each generated Hamiltonian were limited to be between $1$ and $8$ operators long. SISR was applied to each dataset 5 times, recovering the correct Hamiltonian for all 15 trials. This took an average of $4.60 \pm 3.25$ batches, or $225.83 \pm 101.99$ seconds. In 4 out of 15 instances, the Hamiltonian was uncovered after a single epoch. Next, we applied SISR to the same datasets with $(\mu=0, \sigma=0.001)$ of Gaussian noise. Once again, it uncovered the correct Hamiltonian in all trials, requiring an average of $5.00 \pm 3.34$ batches, or $252.76 \pm 109.04$ seconds.

\subsection{Pendulum}

The pendulum is an angular, single-particle dynamical system described by the Hamiltonian
\begin{equation}
    \mathcal{H}(p_\theta,q_\theta) = \frac{p_\theta^2}{2m\ell^2} + mg\ell(1 - \cos(q_\theta))
\end{equation}
where $m$ is mass, $g$ is gravity, and $\ell$ is length.

\begin{figure}[hbt]
\vskip 0.2in
\begin{center}
\centerline{\includegraphics[width=0.90\columnwidth]{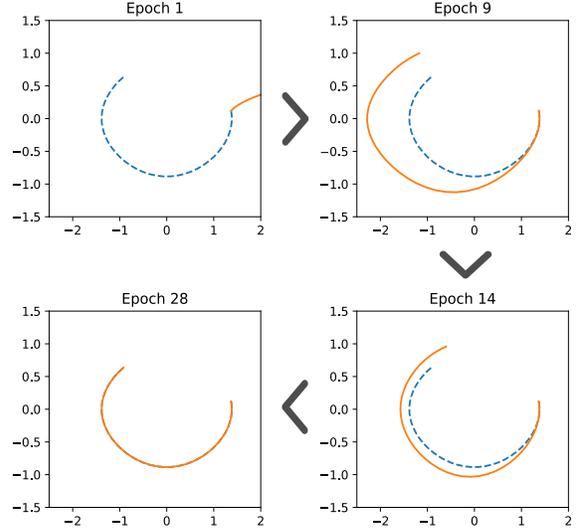}}
\caption{Phase Space $(t=0$ to $t=4)$ for SISR-extracted pendulum Hamiltonian after 1, 9, 14, and 28 epochs (dataset 2 with Gaussian noise). The dashed blue line indicates the ground truth.}
\label{fig:pendulum_trajectory}
\end{center}
\vskip -0.2in
\end{figure}

Using the functional operators $\{+,-,\div,\cdot,\wedge, \cos, \sin\}$, we trained SISR on 3 different datasets of pendulum motion with varying initial conditions, containing $50$ points from $t=0$ to $t=5$. The $T$ and $V$ of each generated Hamiltonian were limited to be between $1$ and $8$ operators long. With no noise present, SISR recovered the correct Hamiltonian in all 15 trials, taking an average of $21.80 \pm 11.18$ batches, or $1049.41 \pm 465.10$ seconds. When $(\mu=0, \sigma=0.005)$ of Gaussian noise was added to each dataset, SISR recovered all correct Hamiltonians in $28.73 \pm 13.68$ batches, or $1117.95 \pm 497.57$ seconds. Hamiltonians generated by SISR for the pendulum problem are depicted in Figure \ref{fig:pendulum_trajectory}.

\subsection{Two-Body Gravitational System}

The two-body gravitational system is a two-dimensional dynamical system described by the Hamiltonian
\begin{equation}
    \mathcal{H}(\mathbf{p}, \mathbf{q}) = \frac{Gm_im_j}{|\mathbf{q}_i - \mathbf{q}_j|} + \frac{\mathbf{p}_i^2}{2m_i} + \frac{\mathbf{p}_j^2}{2m_j}
\end{equation}
with mass $m$ for particles $i$ and $j$. Note that we work explicitly with two-body systems where all masses are equal and $G=1$.

We began by extracting coupling properties via the symplectic neural network procedures described in Section \ref{sec:sympnets}. Using this process, we first determined that particle positions were pairwise coupled, as well as that particle momenta were completely decoupled. Then, we were able to decompose the position coupling into a composite function of Euclidean distance. Finally, we determined that the functions acting on the momenta of each particle are symmetric. The exact numerical results used to guide this search process are available in Appendix \ref{app:sympresults}. Once these coupling constraints are imposed on the RNN, the solution space shrinks dramatically. Rather than learning $V(\mathbf{q})$ and $T(\mathbf{p})$ outright as in our previous experiments, we are effectively learning the pairwise position coupling function between particles, which is itself a function of their Euclidean distance, and the decoupled momentum function acting on each particle.

Using the functional operators $\{+,-,\div,\cdot,\wedge\}$, we trained SISR on 3 different datasets of two-body motion with varying initial conditions, containing $200$ points from $t=0$ to $t=20$. The $T$ and $V$ of each generated Hamiltonian were limited to be between $1$ and $12$ operators long. With no noise present, SISR recovered the correct Hamiltonian for all 15 trials, taking an average of $1.47 \pm 1.30$ batches, or $227.42 \pm 62.94$ seconds. When $(\mu=0, \sigma=0.001)$ of Gaussian noise was added to each dataset, SISR recovered all correct Hamiltonians in $1.80 \pm 1.15$ batches, or $236.79 \pm 55.45$ seconds. Note how powerful our extracted coupling priors are: with them imposed, the two-body Hamiltonian is learned more quickly than a pendulum Hamiltonian.

\begin{table*}[ht]
\caption{Comparison of four methods for extracting governing equations from time series observations. NRMSE refers to normalized RMSE, a reward function between 0 and 1. NRMSE is reported for single-step predictions on the training dataset. We consider an extracted equation "correct" if it is algebraically equivalent with $10^{-2}$ precision in its coefficients.}
\label{table:comparison}
\vskip 0.15in
\begin{center}
\begin{small}
\begin{sc}
\begin{tabular}{cccccccccc}
 & \multicolumn{4}{c}{NRMSE} && \multicolumn{4}{c}{Extracted Correct Equation}\\
\cline{2-5}\cline{7-10}
Problem&SISR&SINDy&SSINN&DSR&&SISR&SINDy&SSINN&DSR\\
\midrule
Harm. Osc. (Clean)    & $1.00 \pm 0.00$ & $0.99 \pm 0.01$ & $1.00 \pm 0.00$ & $1.00 \pm 0.00$ && $\surd$ & $\surd$ & $\surd$ & $\surd$\\
Harm. Osc. (Noisy) & $1.00 \pm 0.00$ & $1.00 \pm 0.00$ & $0.99 \pm 0.00$ & $1.00 \pm 0.00$ && $\surd$ & $\surd$ & $\surd$ & $\surd$\\
Pendulum (Clean)   & $1.00 \pm 0.00$ & $0.99 \pm 0.01$ & $0.92 \pm 0.01$ & $0.99 \pm 0.00$ && $\surd$ & $\times$ & $\times$ & $\surd$\\
Pendulum (Noisy)    & $1.00 \pm 0.00$ & $0.99 \pm 0.01$ & $0.93 \pm 0.02$ & $0.99 \pm 0.00$ && $\surd$ & $\times$ & $\times$ & $\surd$\\
Two-Body (Clean)     & $1.00 \pm 0.00$ & $0.76 \pm 0.18$ & $0.09 \pm 0.07$ & $0.65 \pm 0.23$ && $\surd$ & $\times$ & $\times$ & $\times$\\
Two-Body (Noisy)      & $1.00 \pm 0.00$ & $0.65 \pm 0.15$ & $0.10 \pm 0.03$ & $0.66 \pm 0.23$ && $\surd$ & $\times$ & $\times$ & $\times$\\
Three-Body (Clean)      & $1.00 \pm 0.00$ & $0.72 \pm 0.07$ & $0.18 \pm 0.11$ & $0.77 \pm 0.13$ && $\surd$ & $\times$ & $\times$ & $\times$\\
Three-Body (Noisy)  & $1.00 \pm 0.00$ & $0.69 \pm 0.06$ & $0.16 \pm 0.08$ & $0.75 \pm 0.13$ && $\surd$ & $\times$ & $\times$ & $\times$\\
\bottomrule
\end{tabular}
\end{sc}
\end{small}
\end{center}
\vskip -0.1in
\end{table*}

\subsection{Three-Body Gravitational System}

The three-body gravitational system is a two-dimensional dynamical system described by the Hamiltonian
\begin{equation}
\begin{split}
    \mathcal{H}(\mathbf{p}, \mathbf{q}) = \frac{Gm_im_j}{|\mathbf{q}_i - \mathbf{q}_j|} + \frac{Gm_im_k}{|\mathbf{q}_i - \mathbf{q}_k|} + \frac{Gm_jm_k}{|\mathbf{q}_j - \mathbf{q}_k|} +\\ \frac{\mathbf{p}_i^2}{2m_i} + \frac{\mathbf{p}_j^2}{2m_j} + \frac{\mathbf{p}_k^2}{2m_k}
\end{split}
\end{equation}
with mass $m$ for particles $i$, $j$, and $k$. This system exhibits chaotic motion, meaning that small perturbations in initial conditions diverge exponentially in the future \cite{elegantchaos}.

We followed a nearly identical coupling extraction process to our two-body experiments, identifying that all particles are pairwise coupled as a function of their Euclidean distance, as well as all momenta being completely decoupled and symmetric. These results are detailed in Appendix \ref{app:sympresults}.

Using the functional operators $\{+,-,\div,\cdot,\wedge\}$, we trained SISR on 3 different datasets of three-body motion with $200$ points each. The $T$ and $V$ of each generated Hamiltonian were limited to be between $1$ and $18$ operators long. Five trials were conducted on each dataset. With no noise present, SISR took an average of $2.40 \pm 2.32$ epochs, or $588.92 \pm 238.49$ seconds, to recover the Hamiltonian (all were recovered). With $(\mu=0, \sigma=0.001)$ of Gaussian noise present, SISR required an average of $2.13 \pm 1.77$ epochs, or $575.48 \pm 201.20$ seconds (all were recovered).

\subsection{Comparison to Existing Techniques}

We benchmarked SISR against two other methods for obtaining physical governing equations for data: SINDy and SSINNs. Both of these methods rely upon sparse regressions in user-defined function spaces. They obtained reasonable NRMSE for many problems but only extracted correct governing equations in 2 of 8 instances. Indeed, with so little data, it is easy for sparse-regression oriented methods to overfit to excess terms. Finally, we compared to the vanilla deep symbolic regression model, which we used to generate systems of differential equations instead of Hamiltonians. These results are summarized in Table \ref{table:comparison} with further experimental details available in Appendix \ref{app:comparison}.

\section{Scope and Limitations}

\paragraph{Limited to separable Hamiltonians} SISR uncovers the dynamics of physical systems by generating separable Hamiltonians, but not all physical dynamical systems can be represented in this way. Real-world systems often exhibit energy dissipation that can violate our formalism. Being robust to such dissipation, or knowing when to switch formalisms, may serve as a basis for future work.

\paragraph{Coupling extraction complexity} Our coupling extraction process requires training a large number of black-box symplectic neural networks. Unfortunately, as the physical dynamics of a system become increasingly complicated, these networks must increase in size to maintain predictive accuracy. In doing so, they become more time-consuming to train and further risk overfitting to the system.

\paragraph{Computationally expensive constant optimization} Every expression generated by the LSTM-RNN must have its constants optimized. This means that, over the course of the entire Hamiltonian generation process, many thousands of expressions will need to be optimized. Empirically, this operation accounts for approximately 96\% of the run-time for SISR. Many of these expressions do not fall in the top quantile, meaning that they are not used to trained the RNN. We leave optimizations to this process for future work.

\section{Conclusion}

Here we present Symplectically Integrated Symbolic Regression, a new technique for extracting governing Hamiltonians from time series observations of a physical dynamical system. By applying a model-agnostic coupling extraction approach, we are able to generate priors that greatly benefit our model, allowing it to quickly fit oscillator, pendulum, 2-body and 3-body systems from no more than 200 noisy observations. These results are state-of-the-art in their performance and simultaneous use of small data. Future work may include methods for extracting additional Hamiltonian priors besides coupling, as well as modifications that allow for energy dissipation.

\bibliography{bib}

\begin{thebibliography}{39}
\providecommand{\natexlab}[1]{#1}
\providecommand{\url}[1]{\texttt{#1}}
\expandafter\ifx\csname urlstyle\endcsname\relax
  \providecommand{\doi}[1]{doi: #1}\else
  \providecommand{\doi}{doi: \begingroup \urlstyle{rm}\Url}\fi

\bibitem[Breen et~al.(2020)Breen, Foley, Boekhold, and Zwart]{newtonvsmachine}
Breen, P.~G., Foley, C.~N., Boekhold, T., and Zwart, S.~P.
\newblock Newton versus the machine: solving the chaotic three-body problem
  using deep neural networks.
\newblock \emph{Monthly Notices of the Royal Astronomical Society},
  494\penalty0 (2):\penalty0 2465--2470, 2020.

\bibitem[Brunton et~al.(2020)Brunton, Noack, and Koumoutsakos]{fluids}
Brunton, S., Noack, B., and Koumoutsakos, P.
\newblock Machine learning for fluid mechanics, 2020.
\newblock https://arxiv.org/abs/1905.11075.

\bibitem[Brunton et~al.(2016)Brunton, Proctor, and Kutz]{sindy}
Brunton, S.~L., Proctor, J.~L., and Kutz, J.~N.
\newblock Discovering governing equations from data by sparse identification of
  nonlinear dynamical systems.
\newblock \emph{PNAS}, 113\penalty0 (15):\penalty0 3932--3937, 2016.

\bibitem[Chen et~al.(2019)Chen, Zhang, Arjovsky, and Bottou]{srnn}
Chen, Z., Zhang, J., Arjovsky, M., and Bottou, L.
\newblock Symplectic recurrent neural networks, 2019.
\newblock https://arxiv.org/abs/1909.13334.

\bibitem[Cranmer et~al.(2020{\natexlab{a}})Cranmer, Greydanus, Hoyer,
  Battaglia, Spergel, and Ho]{lagrangian}
Cranmer, M., Greydanus, S., Hoyer, S., Battaglia, P., Spergel, D., and Ho, S.
\newblock Lagrangian neural networks.
\newblock \emph{arXiv preprint arXiv:2003.04630}, 2020{\natexlab{a}}.

\bibitem[Cranmer et~al.(2020{\natexlab{b}})Cranmer, Sanchez-Gonzalez,
  Battaglia, Xu, Cranmer, Spergel, and Ho]{genetic}
Cranmer, M., Sanchez-Gonzalez, A., Battaglia, P., Xu, R., Cranmer, K., Spergel,
  D., and Ho, S.
\newblock {Discovering Symbolic Models from Deep Learning with Inductive
  Biases}.
\newblock 2020{\natexlab{b}}.

\bibitem[DiPietro et~al.(2020)DiPietro, Xiong, and Zhu]{ssinn}
DiPietro, D.~M., Xiong, S., and Zhu, B.
\newblock Sparse symplectically integrated neural networks.
\newblock In \emph{Advances in Neural Information Processing Systems 33}, pp.\
  6074--6085. 2020.

\bibitem[Forest \& Ruth(1990)Forest and Ruth]{fourthorder}
Forest, E. and Ruth, R.~D.
\newblock Fourth-order symplectic integration.
\newblock \emph{Physica {D:} Nonlinear Phenomena}, 43\penalty0 (1):\penalty0
  105--117, 1990.

\bibitem[Greydanus et~al.(2019)Greydanus, Dzamba, and Yosinski]{hnn}
Greydanus, S., Dzamba, M., and Yosinski, J.
\newblock Hamiltonian neural networks.
\newblock In \emph{Advances in Neural Information Processing Systems 32}, pp.\
  15379--15389. 2019.

\bibitem[Haarnoja et~al.(2018)Haarnoja, Zhou, Abbeel, and Levine]{entropy}
Haarnoja, T., Zhou, A., Abbeel, P., and Levine, S.
\newblock Soft actor-critic: Off-policy maximum entropy deep reinforcement
  learning with a stochastic actor.
\newblock In \emph{International conference on machine learning}, pp.\
  1861--1870. PMLR, 2018.

\bibitem[Hand \& Finch(1998)Hand and Finch]{analyticalmechanics}
Hand, L.~N. and Finch, J.~D.
\newblock \emph{Analytical Mechanics}.
\newblock Cambridge University Press, Cambridge, England, 1998.

\bibitem[Jin et~al.(2020)Jin, Zhang, Zhu, Tang, and
  Karniadakis]{jin2020sympnets}
Jin, P., Zhang, Z., Zhu, A., Tang, Y., and Karniadakis, G.~E.
\newblock Sympnets: Intrinsic structure-preserving symplectic networks for
  identifying hamiltonian systems.
\newblock \emph{Neural Networks}, 132:\penalty0 166--179, 2020.

\bibitem[Krizhevsky et~al.(2012)Krizhevsky, Sutskever, and
  Hinton]{classification2}
Krizhevsky, A., Sutskever, I., and Hinton, G.~E.
\newblock Imagenet classification with deep convolutional neural networks.
\newblock In \emph{Advances in Neural Information Processing Systems 25}, pp.\
  1097--1105. 2012.

\bibitem[Liu \& Tegmark(2021)Liu and Tegmark]{poincare}
Liu, Z. and Tegmark, M.
\newblock Machine learning conservation laws from trajectories.
\newblock \emph{Physical Review Letters}, 126\penalty0 (18):\penalty0 180604,
  2021.

\bibitem[Long et~al.(2018)Long, Lu, Ma, and Dong]{pdenet}
Long, Z., Lu, Y., Ma, X., and Dong, B.
\newblock {PDE}-net: Learning {PDE}s from data.
\newblock In \emph{Proceedings of the 35th International Conference on Machine
  Learning}, pp.\  3208--3216, 2018.

\bibitem[Long et~al.(2019)Long, Lu, and Dong]{pdenet2}
Long, Z., Lu, Y., and Dong, B.
\newblock Pde-net 2.0: Learning pdes from data with a numeric-symbolic hybrid
  deep network.
\newblock \emph{Journal of Computational Physics}, 399:\penalty0 108925, 2019.

\bibitem[Lu et~al.(2016)Lu, Ren, and Wang]{genetic2}
Lu, Q., Ren, J., and Wang, Z.
\newblock Using genetic programming with prior formula knowledge to solve
  symbolic regression problem.
\newblock \emph{Computational intelligence and neuroscience}, 2016, 2016.

\bibitem[Mattheakis et~al.(2019)Mattheakis, Protopapas, Sondak, Di~Giovanni,
  and Kaxiras]{mattheakis2019physical}
Mattheakis, M., Protopapas, P., Sondak, D., Di~Giovanni, M., and Kaxiras, E.
\newblock Physical symmetries embedded in neural networks.
\newblock \emph{arXiv preprint arXiv:1904.08991}, 2019.

\bibitem[Morin(2008)]{introclassical}
Morin, D.~J.
\newblock \emph{Introduction to Classical Mechanics: With Problems and
  Solutions}.
\newblock Cambridge University Press, Cambridge, England, 2008.

\bibitem[Mundhenk et~al.(2021)Mundhenk, Landajuela, Glatt, Santiago, Petersen,
  et~al.]{deepsymbolicregressionex2}
Mundhenk, T., Landajuela, M., Glatt, R., Santiago, C., Petersen, B., et~al.
\newblock Symbolic regression via deep reinforcement learning enhanced genetic
  programming seeding.
\newblock \emph{Advances in Neural Information Processing Systems}, 34, 2021.

\bibitem[O’Neill(2009)]{fieldguide}
O’Neill, M.
\newblock Riccardo poli, william b. langdon, nicholas f. mcphee: a field guide
  to genetic programming, 2009.

\bibitem[Petersen et~al.(2019)Petersen, Larma, Mundhenk, Santiago, Kim, and
  Kim]{deepsymbolicregression}
Petersen, B.~K., Larma, M.~L., Mundhenk, T.~N., Santiago, C.~P., Kim, S.~K.,
  and Kim, J.~T.
\newblock Deep symbolic regression: Recovering mathematical expressions from
  data via risk-seeking policy gradients.
\newblock \emph{arXiv preprint arXiv:1912.04871}, 2019.

\bibitem[Qin et~al.(2019)Qin, Wu, and Xiu]{datadriven}
Qin, T., Wu, K., and Xiu, D.
\newblock Data driven governing equations approximation using deep neural
  networks.
\newblock \emph{Journal of Computational Physics}, 395:\penalty0 620--635,
  2019.

\bibitem[Radford et~al.(2019)Radford, Wu, Child, Luan, Amodei, and
  Sutskever]{gpt2}
Radford, A., Wu, J., Child, R., Luan, D., Amodei, D., and Sutskever, I.
\newblock Language models are unsupervised multitask learners.
\newblock 2019.

\bibitem[Raissi \& Karniadakis(2018)Raissi and Karniadakis]{karniadakis}
Raissi, M. and Karniadakis, G.~E.
\newblock Hidden physics models: Machine learning of nonlinear partial
  differential equations.
\newblock \emph{Journal of Computational Physics}, 357:\penalty0 125--141,
  2018.

\bibitem[Raissi et~al.(2020)Raissi, Yazdani, and Karniadakis]{raissi2020hidden}
Raissi, M., Yazdani, A., and Karniadakis, G.~E.
\newblock Hidden fluid mechanics: Learning velocity and pressure fields from
  flow visualizations.
\newblock \emph{Science}, 367\penalty0 (6481):\penalty0 1026--1030, 2020.

\bibitem[Reichl(2016)]{statphysics}
Reichl, L.~E.
\newblock \emph{A Modern Course in Statistical Physics}.
\newblock Wiley-VCH, Weinheim, Germany, 2016.

\bibitem[Saemundsson et~al.(2020)Saemundsson, Terenin, Hofmann, and
  Deisenroth]{saemundsson2020variational}
Saemundsson, S., Terenin, A., Hofmann, K., and Deisenroth, M.~P.
\newblock Variational integrator networks for physically structured embeddings,
  2020.

\bibitem[Sahoo et~al.(2018)Sahoo, Lampert, and Martius]{sahoo2018learning}
Sahoo, S., Lampert, C., and Martius, G.
\newblock Learning equations for extrapolation and control.
\newblock In \emph{International Conference on Machine Learning}, pp.\
  4442--4450. PMLR, 2018.

\bibitem[Sakurai \& Napolitano(2010)Sakurai and Napolitano]{quantmechanics}
Sakurai, J.~J. and Napolitano, J.~J.
\newblock \emph{Modern Quantum Mechanics (2nd Edition)}.
\newblock Pearson, London, England, 2010.

\bibitem[Schmidt \& Lipson(2009)Schmidt and Lipson]{gradient2}
Schmidt, M. and Lipson, H.
\newblock Distilling free-form natural laws from experimental data.
\newblock \emph{Science}, 324\penalty0 (5923):\penalty0 81--85, 2009.

\bibitem[Sprott(2010)]{elegantchaos}
Sprott, J.~C.
\newblock \emph{Elegant Chaos: Algebraically Simple Flows}.
\newblock World Scientific, Singapore, 2010.

\bibitem[Sun et~al.(2019)Sun, Zhang, and Schaeffer]{sun2019neupde}
Sun, Y., Zhang, L., and Schaeffer, H.
\newblock Neupde: Neural network based ordinary and partial differential
  equations for modeling time-dependent data, 2019.

\bibitem[Taylor(2005)]{classmechanics}
Taylor, J.~R.
\newblock \emph{Classical Mechanics}.
\newblock University Science Books, Mill Valley, California, 2005.

\bibitem[Tong et~al.(2021)Tong, Xiong, He, Pan, and Zhu]{tong2021symplectic}
Tong, Y., Xiong, S., He, X., Pan, G., and Zhu, B.
\newblock Symplectic neural networks in taylor series form for hamiltonian
  systems.
\newblock \emph{Journal of Computational Physics}, 437:\penalty0 110325, 2021.

\bibitem[Udrescu \& Tegmark()Udrescu and Tegmark]{sparse3}
Udrescu, S.-M. and Tegmark, M.
\newblock Ai feynman: A physics-inspired method for symbolic regression.
\newblock \emph{Science Advances}, 6\penalty0 (16).

\bibitem[Wang et~al.(2017)Wang, Jiang, Qian, Yang, Li, Zhang, Wang, and
  Tang]{classification1}
Wang, F., Jiang, M., Qian, C., Yang, S., Li, C., Zhang, H., Wang, X., and Tang,
  X.
\newblock Residual attention network for image classification.
\newblock In \emph{Proceedings of the IEEE Conference on Computer Vision and
  Pattern Recognition}, pp.\  3156--3164, 2017.

\bibitem[Xiong et~al.(2020)Xiong, Tong, He, Yang, Yang, and
  Zhu]{xiong2020nonseparable}
Xiong, S., Tong, Y., He, X., Yang, S., Yang, C., and Zhu, B.
\newblock Nonseparable symplectic neural networks.
\newblock \emph{arXiv preprint arXiv:2010.12636}, 2020.

\bibitem[Zhu et~al.(2020)Zhu, Jin, and Tang]{zhu2020deep}
Zhu, A., Jin, P., and Tang, Y.
\newblock Deep hamiltonian networks based on symplectic integrators.
\newblock \emph{arXiv preprint arXiv:2004.13830}, 2020.

\end{thebibliography}
\bibliographystyle{icml2022}

\newpage
\appendix
\onecolumn
\section{Symplectic Integration Pseudocode}\label{app:symp}

We employ the fourth-order symplectic integrator introduced by \cite{fourthorder}, defined in Algorithm 1.

\begin{algorithm}
  \caption{Fourth-Order Symplectic Integrator}
  \label{alg:example}
\begin{algorithmic}
  \STATE {\bfseries Input:} \(\mathbf{q}_{t_i}\), \(\mathbf{p}_{t_i}\), \(t_i\), \(t_{i+1}\), \(\frac{\partial T}{\partial \mathbf{p}}\), \(\frac{\partial V}{\partial \mathbf{q}}\)
  \STATE {\bfseries Output:} \(\mathbf{q}_{t_{i+1}}\), \(\mathbf{p}_{t_{i+1}}\)
  \STATE $h \gets t_{i+1} - t_i$
  \STATE $\mathbf{k}_p \gets \mathbf{p}_{t_i}$
  \STATE $\mathbf{k}_q \gets \mathbf{q}_{t_i}$
  \FOR{$j=1$ {\bfseries to} $4$}
    \STATE $\mathbf{t}_p \gets \mathbf{k}_p$
    \STATE $\mathbf{t}_q \gets \mathbf{k}_q + c_j \cdot \frac{\partial T}{\partial \mathbf{p}}(\mathbf{k}_p) \cdot h$
    \STATE $\mathbf{k}_p \gets \mathbf{t}_p + d_j \cdot \frac{\partial V}{\partial \mathbf{q}}(\mathbf{t}_q) \cdot h$
    \STATE $\mathbf{k}_q \gets \mathbf{t}_q$
  \ENDFOR
  \STATE return $\mathbf{k}_p$, $\mathbf{k}_q$
\end{algorithmic}
\end{algorithm}

This algorithm employs the constants $c_1 = c_4 = \frac{1}{2(2-2^{1/3})}$, $c_2 = c_3 = \frac{1-2^{1/3}}{2(2-2^{1/3})}$, $d_1 = d_3 = \frac{1}{2-2^{1/3}}$, $d_2 = \frac{2^{1/3}}{2-2^{1/3}}$, and $d_4=0$.

\section{Extended Experimental Details}

\subsection{SISR Hyperparameter Selection}\label{app:hyper}

All experiments used the same set of SISR hyperparameters, aside from their expression lengths and functional operators. These hyperparameters were largely obtained from \cite{deepsymbolicregression}, with the exception of mutation rate, number of layers, initial batch size, inner learning rate, inner optimizer, and inner number of epochs. These additional hyperparameters worked well with their initially selected values and did not require any tuning.

All SISR models were trained with a learning rate $\alpha=0.0005$, entropy coefficient $\lambda_H = 0.005$, mutation rate $\mu=0.05$, risk factor $\epsilon=0.95$,  batch size of 500, and initial batch size of 2000. The RNN architecture was a two-layer LSTM with 250 nodes per layer and initial state optimization. It was optimized using Adam. Expressions generated by the RNN were trained for 15 epochs with a learning rate of $0.5$; they were optimized using RMSProp.

\subsection{Initial Conditions}\label{app:initial}

Each experiment used three datasets with different sets of initial conditions. These initial conditions are detailed in Tables \ref{table:harmonicinitial}, \ref{table:penduluminitial}, \ref{table:twobodyinitial}, and \ref{table:threebodyinitial}.

\begin{table}[htb]
\caption{Dataset details for Harmonic Oscillator experiments. These initial conditions were generated randomly.}
\label{table:harmonicinitial}
\vskip 0.15in
\begin{center}
\begin{small}
\begin{sc}
\begin{tabular}{lccccr}
\toprule
Dataset & $m$ & $\omega$ & $q_0$ & $p_0$ \\
\midrule
1    & 1.23 & 1.65 & -0.05 & 0.42 \\
2    & 0.63 & 1.30 & 0.27 & -0.29 \\
3    & 1.69 & 0.83 & 0.06 & -0.54 \\
\bottomrule
\end{tabular}
\end{sc}
\end{small}
\end{center}
\vskip -0.1in
\end{table}

\begin{table}[htb]
\caption{Dataset details for Pendulum experiments. These initial conditions were generated randomly.}
\label{table:penduluminitial}
\vskip 0.15in
\begin{center}
\begin{small}
\begin{sc}
\begin{tabular}{lccccccr}
\toprule
Dataset & $m$ & $\ell$ & $g$ & ${q_\theta}_0$ & ${p_\theta}_0$ \\
\midrule
1    & 0.47 & 1.23 & 1.95 & 1.32 & 0.23 \\
2    & 1.10 & 0.73 & 1.02 & 1.37 & 0.12 \\
3    & 0.68 & 0.33 & 1.59 & 0.87 & 0.15 \\
\bottomrule
\end{tabular}
\end{sc}
\end{small}
\end{center}
\vskip -0.1in
\end{table}

\begin{table}[htb]
\caption{Dataset details for Two-Body Gravitational experiments. These initial conditions were generated by hand to obtain systems exhibiting clear interactions between the bodies and reasonable domains.}
\label{table:twobodyinitial}
\vskip 0.15in
\begin{center}
\begin{small}
\begin{sc}
\begin{tabular}{lccccccr}
\toprule
Dataset & $m_i$ & $m_j$ & ${\mathbf{q}_i}_0$ & ${\mathbf{q}_j}_0$ & ${\mathbf{p}_i}_0$ & ${\mathbf{p}_j}_0$ \\
\midrule
1    & 1 & 1 & (0.00, 0.00) & (1.00, 1.00) & (0.00, 1.00) & (1.00, 0.00)\\
2    & 1 & 1 & (0.00, 0.00) & (0.50, 1.00) & (0.34, 0.30) & (1.00, 0.21)\\
3    & 1 & 1 & (0.00, 0.50) & (0.30, -0.24) & (0.40, -1.00) & (-0.80, -2.10)\\
\bottomrule
\end{tabular}
\end{sc}
\end{small}
\end{center}
\vskip -0.1in
\end{table}

\begin{table}[htb]
\caption{Dataset details for Three-Body Gravitational experiments. These initial conditions were generated by hand to obtain systems exhibiting clear interactions between the bodies and reasonable domains.}
\label{table:threebodyinitial}
\vskip 0.15in
\begin{center}
\begin{small}
\begin{sc}
\begin{tabular}{lcccccccccr}
\toprule
Dataset & $m_i$ & $m_j$ & $m_k$ & ${\mathbf{q}_i}_0$ & ${\mathbf{q}_j}_0$ & ${\mathbf{q}_k}_0$ & ${\mathbf{p}_i}_0$ & ${\mathbf{p}_j}_0$ & ${\mathbf{p}_k}_0$ \\
\midrule
1    & 1 & 1 & 1 & (0.00, 0.00) & (1.00, 1.00) & (-1.00, -1.00) & (0.50, 0.30) & (1.10, -0.40) & (-0.20, 0.50) \\
2    & 1 & 1 & 1 & (0.00, 0.00) & (-3.00, 0.00) & (3.00, 0.00) & (0.00, -1.00) & (0.50, 0.00) & (0.50, 0.00) \\
3    & 1 & 1 & 1 & (0.25, 0.25) & (2.00, 1.50) & (3.00, -2.00) & (0.00, -1.00) & (0.50, -0.15) & (0.80, 0.25) \\
\bottomrule
\end{tabular}
\end{sc}
\end{small}
\end{center}
\vskip -0.1in
\end{table}

\subsection{Extracting Function Properties via Symplectic Neural Networks}\label{app:sympresults}

Using the technique described in Section \ref{sec:sympnets}, we extracted coupling properties for our two-body and three-body experiments. Both systems used the same symplectic network architecture: each block was 8 hidden layers deep with 128 hidden neurons per layer. We used GeLU as our activation function and trained for 3,000 epochs via Adam with a learning rate of $0.0005$. This architecture was largely inspired by the successful three-body prediction network in \citet{newtonvsmachine}, with some slight hand-adjustments made to hyperparameters. In order to reduce overfitting, we trained the network for 10 steps of prediction at a time. To do this, the dataset of 200 points was broken into a set of 180 for training and a set of 10 for testing, such that the set of 10 for testing has no overlap with a 10-step prediction made from any training point. The batch size was set to the entire set of 180 training points.

\subsubsection{Two-Body Results}

Using the architecture described in \ref{app:sympresults}, we began our experiments by attempting to determine the coupling present in the system. We set our maximum tolerable performance decrease to 10\%. Using this approach, our technique correctly deduced for both our clean and noisy datasets that $T(\mathbf{p})$ exhibits pairwise coupling whereas $V(\mathbf{q})$ exhibits complete decoupling; these results are contained in Tables $\ref{table:couplingsearch2bodyclean}$ and $\ref{table:couplingsearch2bodynoisy}$. With these results determined, we began the next search for pairwise position composite functions, again with a maximum tolerable performance decrease of 10\%. This tolerance was met by both Manhattan distance and Euclidean distance composite functions. However, Euclidean distance outperformed Manhattan distance, so it was selected as the composite function. These results are depicted for our clean and noisy datasets in Tables \ref{table:compositesearchclean2body} and \ref{table:compositesearchnoisy2body}. Finally, we assessed whether the completely decoupled momenta functions are symmetric across all quantities; enforcing this symmetry increased performance in both our clean and noisy datasets (Tables \ref{table:symmetry2bodyclean} and Table\ref{table:symmetry2bodynoisy}). Thus, we were able to determine that the $V(\mathbf{q})$ exhibits pairwise coupling as a function of Euclidean distance and that $T(\mathbf{p})$ exhibits complete decoupling with symmetric subfunctions for all quantities.

\subsubsection{Three-Body Results}

Our three-body coupling search preceded similarly to the our two-body search, correctly identifying that $V(\mathbf{q})$ exhibits pairwise coupling as a function of Euclidean distance and that $T(\mathbf{p})$ exhibits complete decoupling with symmetric subfunctions. For this search, however, the tolerable decline was set to 5\%. Additionally, since there are 3 pairs of particles, we performed a recursive backwards elimination with a tolerance of 1\% to see which pairs were necessary--it was determined that all were. Coupling results are contained in Tables \ref{table:couplingsearch3bodyclean} and \ref{table:couplingsearch3bodynoisy} and coupling composite results are contained in Tables \ref{table:compositesearchclean3body} and \ref{table:compositesearchnoisy3body}.

\subsection{Comparison Details}\label{app:comparison}

We compared our method to SINDy, SSINNs, and Vanilla Deep Symbolic Regression. Here we provide some further context on the hyperparameters used for each of these methods.

For SINDy, we employed the standard function library for all tasks except for the pendulum, where we added Fourier terms. The regularization parameter was initially set to $0.10$. As SINDy cannot directly learn Hamiltonians, we instead learned a system of differential equations. If all coefficients were removed for any of the given differential equations, we decreased the regularization parameter in increments of $0.01$ so that this did not happen. If the equations became too long to numerically integrate within $30$ seconds, we raised the regularization parameter in increments of $0.01$ so that this did not happen. We planned on converting these differential equations to Hamiltonians so that we could time-evolve with a symplectic integrator. However, many of the systems of differential equations were not separable, so we instead unrolled their predictions using RK4.

For SSINNs, we employed the 3rd degree polynomial function library for all tasks except for the pendulum, where we added a shiftable sine term. We used the hyperparameters described in \citet{ssinn}, setting a learning rate of $10^{-4}$ with decay and a regularization term of $10^{-3}$. We trained this architecture for 100 epochs.

Finally, we employed vanilla Deep Symbolic Regression. Although not well-suited for physical dynamical systems, we modified this model to produce a series of differential equations, similar to SINDy. Again, these differential equations did not necessarily convert to a separable Hamiltonian, so we trained the model by numerically integrating them with RK4 to obtain a reward. We employed the same architecture and hyperparameters defined by \citet{deepsymbolicregression}, using a 250-node RNN with 0.0005 learning rate, 0.95 risk factor, 0.005 entropy coefficient, and 500 batch size. Then, we trained the model for 200 epochs and optimized with Adam. These hyperparameters worked quite well, as the model achieved reasonable performance for not having any physical priors imposed.

\subsection{Hardware Information}

All SISR models were trained on an Intel i7-9750H CPU, which was found to be comparable in training time to using GPU. Symplectic Neural Networks were trained on an NVIDIA GeForce GTX 1650 GPU.

\begin{table}[htb]
\caption{Coupling detection results for the Two-Body experiment with clean data. Note that pairwise coupling is identical to no coupling being enforced, as there is only one pair.}
\label{table:couplingsearch2bodyclean}
\vskip 0.15in
\begin{center}
\begin{small}
\begin{sc}
\begin{tabular}{lccccr}
\toprule
$T(\mathbf{p})$ & $V(\mathbf{q})$ & NRMSE & Performance Change\\
\midrule
No coupling enforced (baseline) & No coupling enforced (baseline) & $0.90 \pm 0.10$ & N/A (baseline)\\
Complete decoupling enforced & No coupling enforced & $0.56 \pm 0.23$ & $-38.00\% \pm 25.44\%$\\
Dimensional coupling enforced & No coupling enforced & $0.65 \pm 0.24$ & $-28.28\% \pm 26.02\%$\\
Pairwise coupling enforced & No coupling enforced & $0.90 \pm 0.10$ & N/A (baseline) \\
Pairwise coupling enforced & Complete decoupling enforced & $0.84 \pm 0.21$ & $-7.17\% \pm 23.12\%$\\
\bottomrule
\end{tabular}
\end{sc}
\end{small}
\end{center}
\vskip -0.1in
\end{table}

\begin{table}[htb]
\caption{Coupling detection results for the Two-Body experiment with noisy data. Note that pairwise coupling is identical to no coupling being enforced, as there is only one pair.}
\label{table:couplingsearch2bodynoisy}
\vskip 0.15in
\begin{center}
\begin{small}
\begin{sc}
\begin{tabular}{lccccr}
\toprule
$T(\mathbf{p})$ & $V(\mathbf{q})$ & NRMSE & Performance Change\\
\midrule
No coupling enforced (baseline) & No coupling enforced (baseline) & $0.92 \pm 0.08$ & N/A (baseline)\\
Complete decoupling enforced & No coupling enforced & $0.61 \pm 0.22$ & $-34.46\% \pm 23.69\%$\\
Dimensional coupling enforced & No coupling enforced & $0.64 \pm 0.31$ & $-30.27\% \pm 33.82\%$\\
Pairwise coupling enforced & No coupling enforced & $0.92 \pm 0.08$ & N/A (baseline)\\
Pairwise coupling enforced & Complete decoupling enforced & $0.89 \pm 0.11$ & $-3.45\% \pm 11.97\%$\\
\bottomrule
\end{tabular}
\end{sc}
\end{small}
\end{center}
\vskip -0.1in
\end{table}

\begin{table}[htb]
\caption{Coupling composition function results for the Two-Body experiment with clean data.}
\label{table:compositesearchclean2body}
\vskip 0.15in
\begin{center}
\begin{small}
\begin{sc}
\begin{tabular}{lcccr}
\toprule
Composite Function Enforced & NRMSE & Performance Change\\
\midrule
None (baseline) & $0.84 \pm 0.21$ & N/A (baseline)\\
Sum & $0.38 \pm 0.28$ & $-54.72\% \pm 32.99\%$\\
Product & $0.39 \pm 0.28$ & $-54.12\% \pm 33.48\%$\\
Manhattan Distance & $0.93 \pm 0.06$ & $10.91\% \pm 6.73\%$\\
Euclidean Distance & $0.94 \pm 0.05$ & $12.15\% \pm 5.67\%$\\
\bottomrule
\end{tabular}
\end{sc}
\end{small}
\end{center}
\vskip -0.1in
\end{table}

\begin{table}[htb]
\caption{Coupling composition function results for the Two-Body experiment with noisy data.}
\label{table:compositesearchnoisy2body}
\vskip 0.15in
\begin{center}
\begin{small}
\begin{sc}
\begin{tabular}{lcccr}
\toprule
Composite Function Enforced & NRMSE & Performance Change\\
\midrule
None (baseline) & $0.89 \pm 0.11$ & N/A (baseline)\\
Sum & $0.38 \pm 0.28$ & $-57.24\% \pm 31.09\%$\\
Product & $0.39 \pm 0.28$ & $-56.71\% \pm 31.61\%$\\
Manhattan Distance & $0.52 \pm 0.33$ & $-41.81\% \pm 36.95\%$\\
Euclidean Distance & $0.84 \pm 0.20$ & $-5.64\% \pm 22.05\%$\\
\bottomrule
\end{tabular}
\end{sc}
\end{small}
\end{center}
\vskip -0.1in
\end{table}

\begin{table}[htb]
\caption{Coupling symmetry results for the Two-Body experiment with clean data.}
\label{table:symmetry2bodyclean}
\vskip 0.15in
\begin{center}
\begin{small}
\begin{sc}
\begin{tabular}{lcccr}
\toprule
Symmetry Enforced & NRMSE & Performance Change\\
\midrule
None (baseline) & $0.94 \pm 0.05$ & N/A (baseline)\\
Complete symmetry & $0.99 \pm 0.01$ & $5.17\% \pm 0.75\%$\\
\bottomrule
\end{tabular}
\end{sc}
\end{small}
\end{center}
\vskip -0.1in
\end{table}

\begin{table}[htb]
\caption{Coupling symmetry results for the Two-Body experiment with noisy data.}
\label{table:symmetry2bodynoisy}
\vskip 0.15in
\begin{center}
\begin{small}
\begin{sc}
\begin{tabular}{lcccr}
\toprule
Symmetry Enforced & NRMSE & Performance Change\\
\midrule
None (baseline) & $0.84 \pm 0.20$ & N/A (baseline)\\
Complete symmetry & $0.91 \pm 0.10$ & $8.39\% \pm 11.92\%$\\
\bottomrule
\end{tabular}
\end{sc}
\end{small}
\end{center}
\vskip -0.1in
\end{table}

\begin{table}[htb]
\caption{Coupling detection results for the Three-Body experiment with clean data. Note that when performing the recursive backwards elimination, the baseline is shifted to the same coupling type being employed, except without the specific pair eliminated.}
\label{table:couplingsearch3bodyclean}
\vskip 0.15in
\begin{center}
\begin{small}
\begin{sc}
\begin{tabular}{lccccr}
\toprule
$T(\mathbf{p})$ & $V(\mathbf{q})$ & NRMSE & Performance Change\\
\midrule
No coupling enforced (baseline) & No coupling enforced (baseline) & $0.92 \pm 0.08$ & N/A (baseline)\\
Complete decoupling enforced & No coupling enforced & $0.62 \pm 0.20$ & $-32.65\% \pm 22.07\%$\\
Dimensional coupling enforced & No coupling enforced & $0.76 \pm 0.21$ & $-17.72\% \pm 22.61\%$\\
Pairwise coupling enforced & No coupling enforced & $0.98 \pm 0.01$ & $6.54\% \pm 1.60\%$\\
Pairwise coupling (i,j and j,k) & No coupling enforced & $0.95 \pm 0.05$ & $-2.8\% \pm 5.46\%$\\
Pairwise coupling (i,j and i,k) & No coupling enforced & $0.94 \pm 0.07$ & $-3.8\% \pm 7.16\%$\\
Pairwise coupling (i,k and j,k) & No coupling enforced & $0.89 \pm 0.10$ & $-9.7\% \pm 10.02\%$\\
Pairwise coupling enforced & Complete decoupling enforced & $0.88 \pm 0.11$ & $-4.57\% \pm 11.4\%$\\
\bottomrule
\end{tabular}
\end{sc}
\end{small}
\end{center}
\vskip -0.1in
\end{table}

\begin{table}[htb]
\caption{Coupling detection results for the Three-Body experiment with noisy data. Note that when performing the recursive backwards elimination, the baseline is shifted to the same coupling type being employed, except without the specific pair eliminated.}
\label{table:couplingsearch3bodynoisy}
\vskip 0.15in
\begin{center}
\begin{small}
\begin{sc}
\begin{tabular}{lccccr}
\toprule
$T(\mathbf{p})$ & $V(\mathbf{q})$ & NRMSE & Performance Change\\
\midrule
No coupling enforced (baseline) & No coupling enforced (baseline) & $0.92 \pm 0.12$ & N/A (baseline)\\
Complete decoupling enforced & No coupling enforced & $0.67 \pm 0.15$ & $-26.58\% \pm 15.96\%$\\
Dimensional coupling enforced & No coupling enforced & $0.83 \pm 0.15$ & $-9.52\% \pm 16.47\%$\\
Pairwise coupling enforced & No coupling enforced & $0.98 \pm 0.01$ & $7.15\% \pm 1.47\%$\\
Pairwise coupling (i,j and j,k) & No coupling enforced & $0.96 \pm 0.04$ & $-1.86\% \pm 3.78\%$\\
Pairwise coupling (i,j and i,k) & No coupling enforced & $0.94 \pm 0.08$ & $-4.16\% \pm 7.88\%$\\
Pairwise coupling (i,k and j,k) & No coupling enforced & $0.92 \pm 0.07$ & $-6.34\% \pm 6.79\%$\\
Pairwise coupling enforced & Complete decoupling enforced & $0.98 \pm 0.02$ & $6.4\% \pm 2.35\%$\\
\bottomrule
\end{tabular}
\end{sc}
\end{small}
\end{center}
\vskip -0.1in
\end{table}

\begin{table}[htb]
\caption{Coupling composition function results for the Three-Body experiment with clean data.}
\label{table:compositesearchclean3body}
\vskip 0.15in
\begin{center}
\begin{small}
\begin{sc}
\begin{tabular}{lcccr}
\toprule
Composite Function Enforced & NRMSE & Performance Change\\
\midrule
None (baseline) & $0.88 \pm 0.11$ & N/A (baseline)\\
Sum & $0.56 \pm 0.17$ & $-35.92\% \pm 19.74\%$\\
Product & $0.52 \pm 0.21$ & $-41.01\% \pm 23.78\%$\\
Manhattan Distance & $0.70 \pm 0.21$ & $-20.74\% \pm 23.71\%$\\
Euclidean Distance & $0.99 \pm 0.01$ & $12.51\% \pm 1.06\%$\\
\bottomrule
\end{tabular}
\end{sc}
\end{small}
\end{center}
\vskip -0.1in
\end{table}

\begin{table}[htb]
\caption{Coupling composition function results for the Three-Body experiment with noisy data.}
\label{table:compositesearchnoisy3body}
\vskip 0.15in
\begin{center}
\begin{small}
\begin{sc}
\begin{tabular}{lcccr}
\toprule
Composite Function Enforced & NRMSE & Performance Change\\
\midrule
None (baseline) & $0.98 \pm 0.02$ & N/A (baseline)\\
Sum & $0.56 \pm 0.17$ & $-42.45\% \pm 17.75\%$\\
Product & $0.52 \pm 0.21$ & $-47.31\% \pm 21.78\%$\\
Manhattan Distance & $0.70 \pm 0.21$ & $-28.74\% \pm 21.46\%$\\
Euclidean Distance & $0.99 \pm 0.01$ & $0.57\% \pm 0.70\%$\\
\bottomrule
\end{tabular}
\end{sc}
\end{small}
\end{center}
\vskip -0.1in
\end{table}


\end{document}